\documentclass[11pt,a4paper]{article}
\usepackage[hyperref]{eacl2021}
\usepackage{times}
\usepackage{latexsym}

\usepackage{microtype}

\aclfinalcopy 
\def\aclpaperid{428} 

\usepackage{amsmath,amsfonts,bm}

\def\eqref#1{equation~\ref{#1}}

\def\1{\bm{1}}

\def\vx{{\bm{x}}}
\def\vy{{\bm{y}}}

\DeclareMathAlphabet{\mathsfit}{\encodingdefault}{\sfdefault}{m}{sl}
\SetMathAlphabet{\mathsfit}{bold}{\encodingdefault}{\sfdefault}{bx}{n}

\newcommand{\Ls}{\mathcal{L}}

\usepackage{mathabx}
\usepackage{subfig}
\usepackage{float}
\usepackage{times}
\usepackage{soul}
\usepackage{url}
\usepackage{hyperref}
\usepackage[utf8]{inputenc}
\usepackage{caption}
\usepackage{graphicx}
\usepackage{multirow}
\usepackage{amsmath}
\usepackage{amsthm}
\usepackage{booktabs}
\usepackage{algorithm}
\usepackage{algorithmic}

\title{Analyzing the Forgetting Problem in Pretrain-Finetuning \\ of Open-domain  Dialogue Response Models}

\author{Tianxing He \\
  MIT \\
  \texttt{tianxing@mit.edu} \\\And
  Jun Liu \\
  Facebook \\
  \texttt{junliu@fb.com} \\ \And 
  Kyunghyun Cho \\
  New York University \\
  \texttt{kyunghyun.cho@nyu.edu} \\ \AND
  Myle Ott \\
  Facebook \\
  \texttt{myleott@fb.com} \\ \And
  Bing Liu \\
  Facebook \\
  \texttt{bingl@fb.com} \\ \And
  James Glass \\
  MIT\\
  \texttt{glass@mit.edu} \\ \And
  Fuchun Peng \\
  Facebook \\
  \texttt{fuchunpeng@fb.com} \\
  }

\date{}

\begin{document}
\maketitle
\begin{abstract}
In this work, we study how the finetuning stage in the pretrain-finetune framework changes the behavior of a pretrained neural language generator. We focus on the transformer encoder-decoder model for the open-domain dialogue response generation task. Our major finding is that after standard finetuning, the model \textit{forgets} some of the important language generation skills acquired during large-scale pretraining. We demonstrate the forgetting phenomenon through a set of detailed behavior analysis from the perspectives of knowledge transfer, context sensitivity, and function space projection. As a preliminary attempt to alleviate the forgetting problem, we propose an intuitive finetuning strategy named ``mix-review''. We find that mix-review effectively regularizes the finetuning process, and the forgetting problem is alleviated to some extent. Finally, we discuss interesting behavior of the resulting dialogue model and its implications.
\end{abstract}

\section{Introduction}
\label{sec:intro}

Large-scale unsupervised pretraining \cite{elmo18peters,jacob18bert,song2019mass,xlnet19zhilin,yinhan19roberta} has recently been shown to greatly boost the performance of natural language processing (NLP) models. On a high level, the pretrain-finetune framework can be viewed as a simple two-stage procedure: (1) Use large-scale unsupervised text data to pretrain the model; (2) Use target task data to finetune the model.

Recently, multiple works \citep{radford2019language,jiang2020know,roberts2020knowledge,talmor2019olmpics} have reported that pretrained language models (LM) have implicitly stored large amounts of ``world knowledge'' in its parameters, and are able to answer common-sense questions. While these studies are encouraging, during the finetuning stage the model is usually trained on a dataset that is very different from the pretraining data, which leads to the potential danger that the model could \textbf{forget} precious skills gained during pretraining. This is an important question for open-domain dialogue response generation, which is the focus of our work, because the knowledge acquired during pretraining can greatly help make the dialogue interaction more engaging or informative.



\begin{figure}[t]
    \centering
    \includegraphics[width=0.35\textwidth]{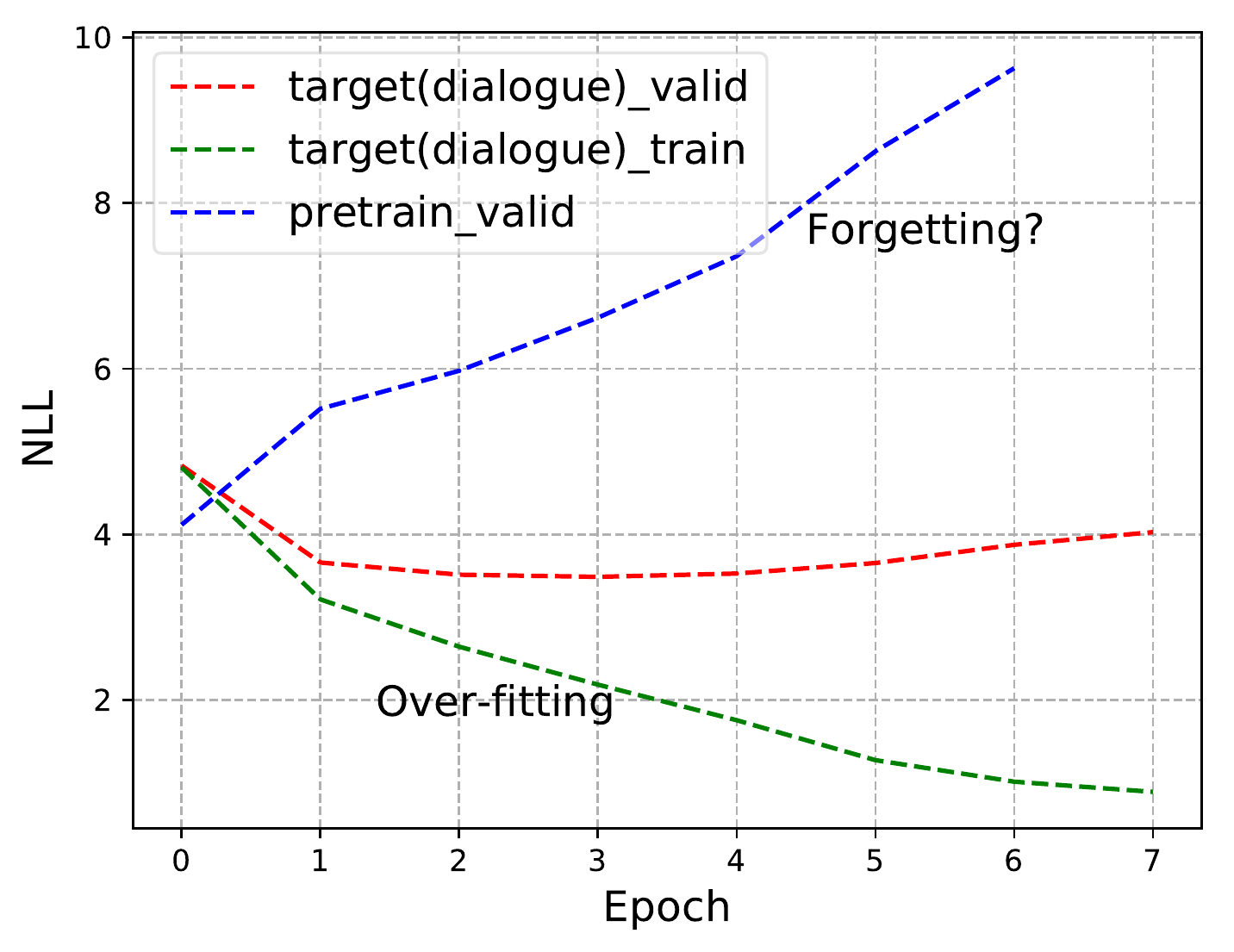}
    \caption{During finetuning, the model's performance on the pretraining data drastically degrades.}
    \label{fig:intro_forget}
\end{figure}

In Figure \ref{fig:intro_forget}, we show that during finetuning, the model's performance on the pretraining data drastically degrades. While this drop is concerning, it does not necessarily mean that the skills from pretrained model are not well transferred to the end dialogue task, because the model should be evaluated in a dialogue setting.

To better answer the question about how finetuning changes the pretrained model's behavior, in this work we conduct a set of behavior analysis from the perspectives of knowledge transfer, context sensitivity, and function space projection. Our major finding is that in the finetuning stage, data separation causes the model to \textit{forget} some of the important language generation skills acquired during pretraining. We also show that the forgetting problem can be alleviated by mixing pretraining and target-task data during finetuning.
\section{Model Formulation}
\label{sec:model}

In this work we study the pretrain-finetune framework from the viewpoint of neural language generation (NLG). In particular, we focus on the open-domain dialogue response task, for the following reasons: 
(1) There is high similarity between the target dialogue response task (conditional NLG) and the pretraining language modeling (LM) objective, so we expect that language generation skills learnt during pretraining can be well transferred to the down-stream target task. (2) The sequence-to-sequence (seq2seq) nature of the model allows us to characterize the model's generation behavior in various ways (e.g., context sensitivity).

End-to-end dialogue response generation \cite{diversityjiwei16} can be formulated as a sequence-to-sequence (seq2seq) task: Given a dialogue context (previous utterances), the model is asked to generate a high-quality response. In this work we adopt the encoder-decoder model architecture \cite{ilya14seq,cho-al-emnlp14}, which is widely used in NLG applications like dialogue response generation \cite{diversityjiwei16}, machine translation \cite{thang-att-mt-15}, etc. In particular, we use the transformer model \cite{tfattention17Vaswani}, which has currently become the most popular encoder-decoder model architecture \cite{trend17tom}. We use the same configuration as \cite{tfattention17Vaswani}, which has 6 encoder/decoder layers, 16 attention heads, with an embedding dimension of 1024 and a feed-forward dimension of 4096.

During standard finetuning, the Adam optimizer \cite{adam14kingma} is used to minimize the negative log-likelihood (NLL) of the reference target sentence $\vy$ given the input context $\vx$ in the data distribution (denoted as $P_{data}$):
\begin{equation}
\label{eq:mle}
\begin{split}
 \Ls_{\text{finetune}}  (P_{data};\theta)  = E_{(\vx, \vy) \sim P_{data}} (-\log P_{\theta}(\vy|\vx)) \\
  = E_{(\vx, \vy) \sim P_{data}} (- \sum^m_{t=1} \log P_{\theta}(y_t|\vy_{<t}, \vx)), 
\end{split}
\end{equation}
where $\vy_{<t}$ refers to $\{y_0,y_1,...,y_{t-1}\}$, in which $y_0$ is set to a begin-of-sentence token \texttt{<BOS>}, and $y_m$ is a end-of-sentence token \texttt{<EOS>}. In the dialogue response setting, the input $\vx$ is a concatenation of previous utterances. We truncate the length of $\vx$ to be at most 128 words, which typically includes around 6 previous utterances. 


Given a trained seq2seq model, to generate a response for some contextual input, one needs to choose a decoding or sampling method. Recent research \cite{curious19ari,radford2019language,fan2018-storyhierarchical} has shown that a strategy called top-$k$ sampling, in which the next word is sampled from the top $k$ most probable choices, is a better choice than the traditional beam-search decoding, due to better diversity. Our preliminary experiments (Appendix \ref{appsec:beamsearch}) have also verified this claim in the open-domain dialogue response setting. As a result, in this work, unless otherwise mentioned, we use top-$k$ sampling as the default decoding method. In particular, we set $k$ to 30 (we find it to work well in preliminary experiments).


\section{The Pretrain-Finetune Framework}
\label{sec:pretrainfinetune}

In this section we first review the pretrain-finetune framework for encoder-decoder models. We discuss the language generation skills the model can acquire during pretraining, and more importantly, how we check whether the skills are ``forgotten'' during finetuning. Finally, as a preliminary attempt to alleviate the forgetting problem, we propose the mix-review finetuning strategy.

\subsection{Pretraining}
\label{sec:pretrain}

\begin{table}
\small
\begin{tabular}{l}
\hline
\textbf{Dialogue} \\
\textbf{Input:} what did you do yesterday ? \texttt{<eou>} \\ i watched the avengers movie . \\
\textbf{Output:} wow ! i am crazy about iron man ! \\
\hline
\textbf{Next-sentence Pretraining} \\
\textbf{Input:} the avengers are super hot currently . \texttt{<eou>} \\ the next movie will be on in April .   \\
\textbf{Output:} fans are talking about iron man on the internet . \\
\hline
\textbf{MASS Pretraining} \\
\textbf{Input:} fans are talking about \texttt{<MASK> <MASK> <MASK>} \\ will do on the internet . \\
\textbf{Output:} what iron man \\
\hline
\end{tabular}
\caption{Illustrations of input-output pairs for typical dialogue response training, next-sentence pretraining, or MASS pretraining.}
\label{tab:illustrate-ns}
\end{table}

In this work, we consider pretraining the seq2seq model using large-scale unsupervised text data, and afterwards finetuning it using target dialogue data. We compare two representative strategies: next-sentence (NS) pretraining and masked sequence-to-sequence (MASS) pretraining \cite{song2019mass}. Next-sentence pretraining is a natural extension of GPT-style LM training \cite{radford2019language,ryan15skip} for encoder-decoder models. For every sentence in a given training document, we set the previous sentences as the contextual input, and ask the model to generate the next sentence. We omit the formulation of NS because it is very similar to Equation (\ref{eq:mle}).

Masked sequence-to-sequence pretraining (MASS) can be regarded as an extension of the ``BERT'' \cite{jacob18bert} pretraining for encoder-decoder models. For each sentence, a random segment of the sentence is masked, and the model is trained to generate the masked words on the decoder side. We refer readers to \cite{song2019mass} for more details.


In Table \ref{tab:illustrate-ns}, we illustrate the similarity between NS pretraining and typical dialogue response training. Compared to NS pretraining, MASS has the disadvantage that it focuses on one single sentence at a time.  However, the context of multiple previous sentences are very important for dialogue response generation.

\subsection{Analyzing the Forgetting Problem}
\label{sec:analyze_forgetting}

Although recently a number of pretraining strategies \cite{elmo18peters,jacob18bert,song2019mass,xlnet19zhilin,yinhan19roberta} have been proposed for various NLP tasks, the finetuning stage remains simple and straightforward: simply finetune all parameters with a relatively small learning rate.

\begin{figure}[h]
\subfloat[Mix-review]{\label{fig:forgetting_mixreview_nll}\includegraphics[width=0.53\linewidth]{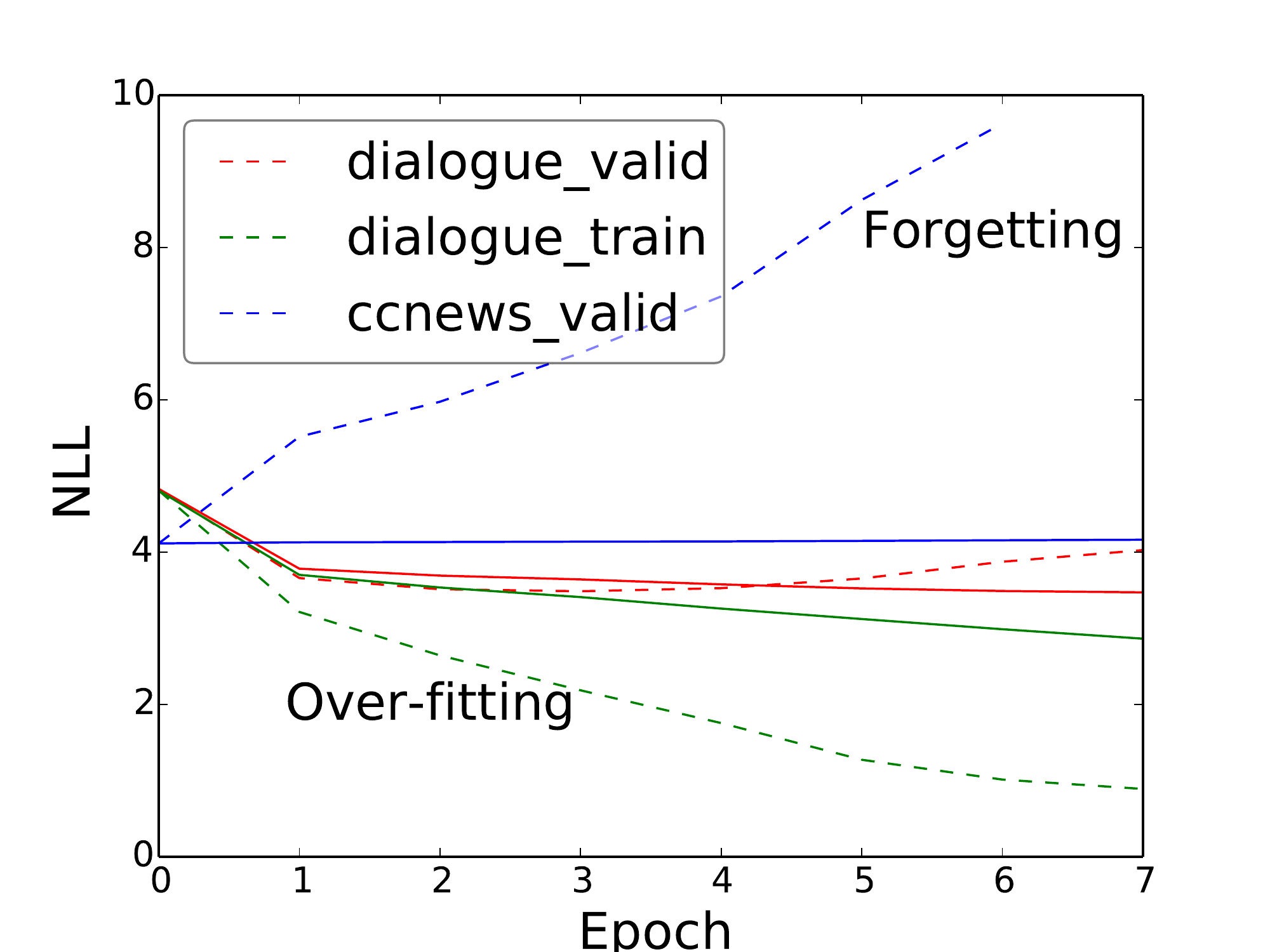}}
\subfloat[WD($\theta_\text{pre}$)]{\label{fig:forgetting_wdecay_nll}\includegraphics[width=0.53\linewidth]{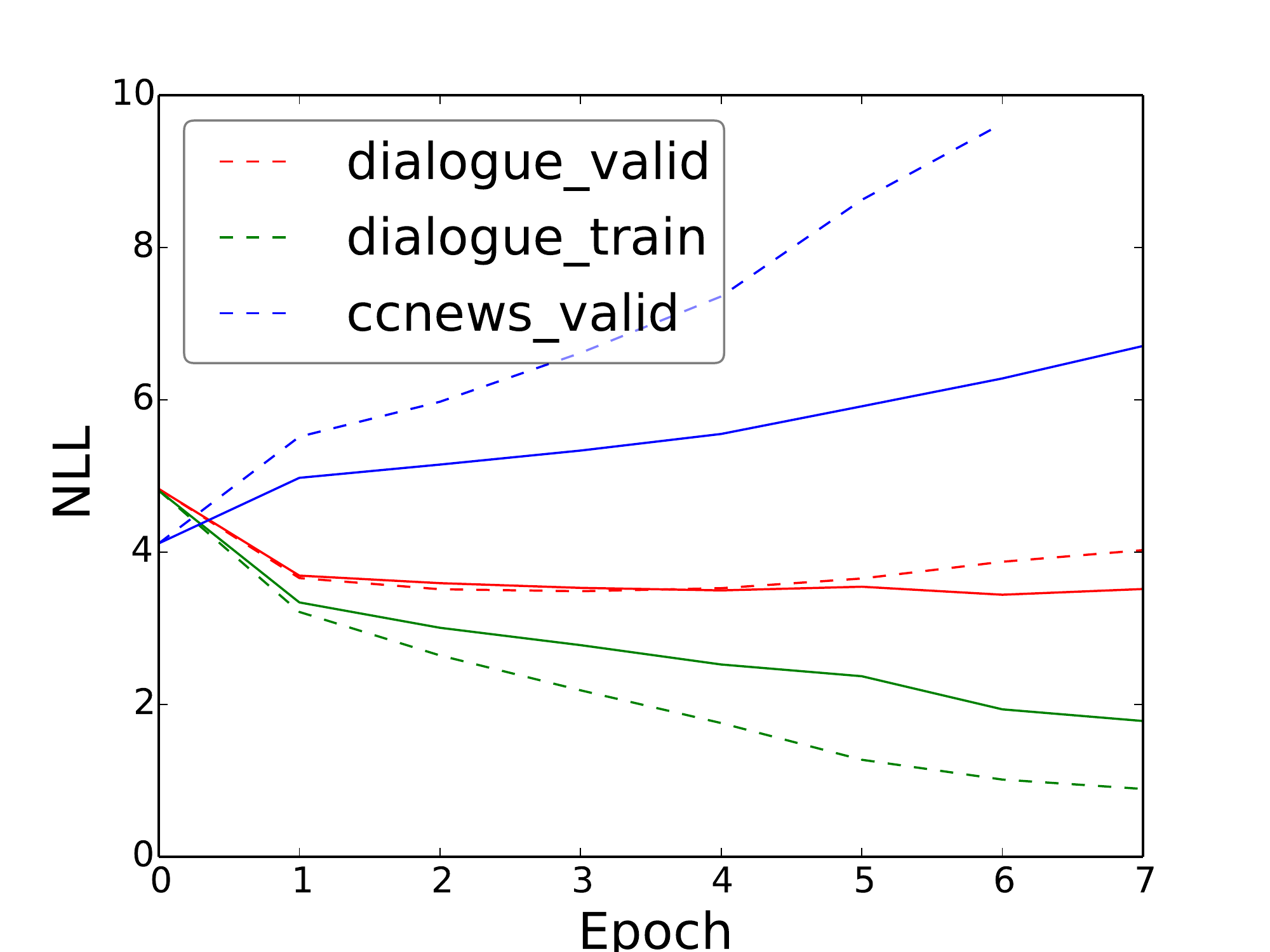}}
\caption{Model's performance on different evaluation sets during the finetuning stage, for the Dailydialog dataset (to be described in Section \ref{sec:datasets}). The dotted lines represent the original finetuning process, while the solid lines represent when mix-review or WD($\theta_\text{pre}$) is applied.}
  \label{fig:forgetting_nll}
\end{figure}


In Figure \ref{fig:forgetting_mixreview_nll}, we show (with the dotted lines) the model's negative log-likelihood (NLL) on different evaluation sets during the finetuning stage. We identify two potential issues during finetuning: (1) \textbf{Over-fitting:} The gap between training-set NLL and validation-set NLL increases quickly. (2) \textbf{Forgetting:} The performance on the pretraining CCNEWS data (to be described in Section \ref{sec:datasets}) drops drastically. Note that the forgetting phenomenon here is not necessarily ``catastrophic'' as in the sequential learning case \cite{pesudo18craig,Robins95catastrophicforgetting}, because the goal is to achieve the best performance on the target dialogue dataset, and the model does not need to maintain fidelity to the pretraining data. However, \textbf{it leads us to question whether the model has lost some important skills learned during pretraining}.

In this work we analyze two important generation capabilities that the model can acquire in the pretraining stage, and will be useful for the target dialogue setting. One is the \textit{acquisition of knowledge}: the large-scale pretraining text data contains a large amount of knowledge, and can be used to make dialogue responses more informative and engaging (e.g., the model can learn about the ``Avengers'' movie, and use it as a topic). To quantify how knowledgeable the finetuned model is, we prepare a set of knowledge terms such as iphone, pokemon, etc., and the corresponding reference description. We then query the model about these knowledge terms, and compare its output against the reference. 
We also conduct multi-turn human evaluation in the setting of knowledgeable conversations. More details will be given in Section \ref{sec:behav_knowledge}.

The other ability is the \textit{utilization of contextual input}: as shown by \cite{hisotrydia19Chinnadhurai}, the current open-domain dialogue models (without pretraining) are insensitive to contextual input, which gives rise to the generic response problem \cite{diversityjiwei16}. In our preliminary experiments with NS pretraining, we find that similarly to the GPT model \cite{radford2019language} the pretrained model has the ability to generate closely related responses given the previous sentences as input. Ideally during finetuning, the model can transfer this skill to the target dialogue task.  To quantify the model's sensitivity to context, following \cite{hisotrydia19Chinnadhurai}, we add noise to the input, and measure the relative drop in perplexity. More details will be given in Section \ref{sec:bea_sens}.

\subsection{The Mix-review Finetuning Strategy}
\label{sec:forget_mixreview}

As a preliminary attempt to alleviate the forgetting problem, we propose a finetuning strategy named ``mix-review (MR)'': For each finetuning epoch, we mix the target dialogue data with a random subset of the pretraining data. This process introduces two hyper-parameters: \textit{mix-ratio}, which controls how much pretraining data is mixed, and \textit{mix-decay}, which decays mix-ratio by each epoch. For example, assume the target dialogue training set has 100k utterances, mix-ratio$=$4 and mix-decay$=$0.9, then in the first epoch of mix-review finetuning, 400k pretraining utterances will be mixed in, and for the second epoch the amount will be reduced to 360k utterances, etc.

We formulate the mix-review objective as below:
\begin{equation}
\label{eq:mix_review}
\begin{split}
\Ls_{\text{mix-review}} = \Ls_{\text{finetune}} (P_{\text{target-data}};\theta) + \\
 \text{mix-ratio} \cdot \Ls_{\text{pretrain}} (P_{\text{pretrain-data}};\theta).
 \end{split}
\end{equation}
 Note that the augmented mixing term can be viewed as a regularization term.

We tune the hyper-parameters (mix-ratio and mix-decay) in the grid of $\{1, 2, 4, 8, 16\} \bigtimes \{1, 0.9, 0.8, 0.7, 0.6, 0.5\}$ (using the same learning rate and other hyper-parameters with standard finetuning), and report with the best model based on the perplexity (PPL) performance on the validation set of the target task. We find that the performance gain of mix-review is not sensitive to hyper-parameter tuning: a small mix-ratio of 4 typically works well, which means the computational cost of mix-review is comparable to standard finetuning.

In Figure \ref{fig:forgetting_mixreview_nll}, we show the loss curve for mix-review finetuning with a mix-ratio of 4 and a mix-decay of 0.7. We observe that the performance on the pretraining CCNEWS data is preserved, which strongly supports the motivation of mix-review. Furthermore, we observe a regularization effect from mix-review (narrowing the gap between training and testing performance).

We compare mix-review with the $L_2$ regularization (weight decay) toward the pretrained parameters $\theta_\text{pre}$ \cite{Kirkpatrick2016OvercomingCF}. We denote it as WD($\theta_\text{pre}$) and formulate it as follows:
\begin{equation}
\label{eq:wd_pre}
 \Ls_{\text{finetune}} (P_{\text{target-data}};\theta) + \lambda \cdot \left\lVert \theta - \theta_\text{pre} \right\rVert^2_2.
\end{equation}
In our experiments, we tune $\lambda$ in the set \{$10^{-1}$,$10^{-2}$,$10^{-3}$,$10^{-4}$,$10^{-5}$\} and report with the best model based on PPL on the validation set. 

In Figure \ref{fig:forgetting_wdecay_nll} we show the loss curve for WD($\theta_\text{pre}$) with $\lambda=0.1$. We observe that WD($\theta_\text{pre}$) also has a regularization effect, but it is not as strong as mix-review.

Additionally, we tried the following two basic regularization techniques: (1) Increase the rate of dropout; (2) Freeze the bottom layers of the model during finetuning. However, these two techniques show little or no improvement. The reason could be that the transformer is already a well-tuned model (e.g., it features dropout and layer normalization).




\section{Datasets and Implementation Details}
\label{sec:data_implement}


\subsection{Datasets}
\label{sec:datasets}

For pretraining, we use the large-scale CCNEWS data \cite{anton19realfake} which is a de-duplicated subset of the English portion of the CommonCrawl news dataset\footnote{
\url{http://commoncrawl.org/2016/10/news-dataset-available}
}.
The dataset contains news articles published worldwide between September 2016 and February 2019. It has in total around 1 billion sentences or 27 billion words. To be able to complete experiments in a reasonable amount of time, we use the first 10 percent of the CCNEWS data for pretraining, which contains 100 million sentences and 2.7 billion words. 

For finetuning, three open-domain conversational dialogue datasets are used: Dailydialog (1.3 million words) \cite{dailydialog17yanran}, Switchboard (1.2 million words), and Cornell Movie \cite{cornell11cristian} (4.5 million words). To save space, we defer the details of the data-sets to Appendix \ref{appsec:dataset}.

To construct the vocabulary, we learn codes of Byte Pair Encoding (BPE) \cite{bpe16sennrich} from the CCNEWS-100m data with 50k merges. This results in a vocabulary of size 62k. We then apply the same BPE codes to all target dialogue datasets.  

\subsection{Implementation}

Our code is based on the Fairseq toolkit \cite{ott2019fairseq}. The Adam optimizer \cite{adam14kingma} is used for all experiments. For pretraining of both MASS and NS, we use a mini-batch size of 2048, with the learning rate (LR) set to 0.0001. Following \cite{tfattention17Vaswani}, the ``inverse square root'' LR scheduler with a warm-up stage is used. Pretraining is conducted on 32 GPUs and half-precision (float16) speed-up is used. For both MASS and NS, we stop the pretraining after the CCNEWS data is swept 20 times.  For all our experiments, a dropout rate of 0.1 is applied to the transformer model. We follow \citet{song2019mass} for the recommended hyper-parameter setting of MASS (e.g., how to select the mask span).

Finetuning is done on 2 GPUs without float16 speed-up. The learning rate is halved when the PPL on the validation set does not improve. In almost all finetuning experiments over-fitting is observed, and we do an early-stop when performance on the validation set starts to deteriorate. We tune the learning rate from \{$10^{-3}$,$10^{-4}$,$10^{-5}$\}, and report the best model based on validation set perplexity.


\section{Experiment Results}
\label{sec:exp}


In this section, we conduct a set of detailed behavior analysis, characterising how different training strategies change the model's behavior. In particular, we aim to answer the crucial question about whether the model forgets precious language generation skills during standard finetuning, and whether mix-review helps the model remember the skills.


\begin{table}
\small
\begin{center}
\begin{tabular}{|l|c|c|c|}
\hline
\multirow{2}{*}{\textbf{Training}} & \multicolumn{3}{c|}{ \textbf{Test-PPL}}        \\ \cline{2-4} 
                          & {\bf DD} & {\bf SB} & {\bf CM} \\ \hline
Baseline(from scratch)    &      24.83  & 51.14 & 49.48
          \\ \hline
MASS+finetune &      12.78    &     28.41
&      30.25    \\ \hline
NS+finetune   &   11.54  &    26.37 &     28.06  \\ 
\hline
NS+WD($\theta_\text{pre}$) & 11.19 & 26.25 & 27.80 \\
\hline
NS+mix-review & {\bf 11.07} & {\bf 25.92} & {\bf 27.54} \\
\hline
\end{tabular}
\end{center}
\caption{Perplexity on test set for different training process on the three dialogue datasets. The datasets are Dailydialogue (DD), Switchboard (SB), and Cornell Movie (CM).}
\label{tab:pplamt}
\end{table}

We first present perplexity results for different finetuning methods in Table \ref{tab:pplamt}. We observe the big improvement in perplexity (larger than 40\%) for the pretrained models comparing to the baseline models trained from scratch. Comparing to MASS, the NS pretraining has more than 7\% relative improvement. This confirms our earlier discussion that the model pretrained by NS better utilizes contextual input (which is further verified in Section \ref{sec:bea_sens}). Based on this observation, we focus our analysis below on the NS pretraining. 

Comparing to standard finetuning, mix-review further gives solid improvement. The gain is due to its strong regularization effect (which we study in the next three sections). However, the performance gap between mix-review and WD($\theta_\text{pre}$) is not significant. We believe the reason is that the benefit (e.g., knowledge transfer) from alleviate the forgetting problem is not be well demonstrated in single-turn response evaluation, because the context is limited to the narrow scope of the specific datasets. We address this concern with multi-turn human evaluation in the next section. 

\subsection{Behavior Analysis: Knowledge Transfer}
\label{sec:behav_knowledge}

As argued in Section \ref{sec:pretrain}, ideally the model can acquire common-sense (or world) knowledge from the large-scale pretraining data, which will be useful for the downstream open-domain dialogue task. In this section, we design a process to quantify how much knowledge the model has, and use it to monitor how the pretrain-finetune framework changes the model's behavior. 

Since the pretraining CCNEWS data is in the public news domain, we expect the model to have knowledge about ``big news''. So, we utilize the Google trend data of the year 2016\footnote{\url{https://www.google.com/intl/en-US/trends/2016records/}},
which contains 365 trending terms (e.g., iPhone 7, Deadpool), and its corresponding description.

\begin{table*}[]
\centering
\small
\begin{tabular}{l|l}
\hline
\textbf{News-style Triggers} & \textbf{Dialogue-style Triggers} \\
\hline
now, some opinions about X .  \hspace{1.3cm}  & what you do think about X ? \\
let me tell you about X . & please tell me about X . \\
here's some news about X . & do you have news about X ? \\
\hline
\multicolumn{2}{l}{\textbf{Reference Description:}  Pokemon first took the world by storm in the mid-90s, doing so once  }   \\ 
\multicolumn{2}{l}{again this year with the release of Pokemon Go. } \\
\multicolumn{2}{l}{ \textbf{NS Pretrained:} the game , titled pokemon go : pocket camp , can be played in person ... } \\
\multicolumn{2}{l}{ \textbf{Standard Finetuned:} it 's a new game that can be played with kids .} \\
\multicolumn{2}{l}{ \textbf{WD($\theta_\text{pre}$):} pokemon go , it 's a type of game that only exists in the us .} \\
\multicolumn{2}{l}{ \textbf{Mix-review:} pokemon go is a popular mobile game , where you 're expected to catch pokemon .} \\
\hline
\multicolumn{2}{l}{\textbf{Reference Description:} Deadpool: The wisecracking antihero, played by Ryan Reynolds in a }   \\ 
\multicolumn{2}{l}{movie of the same name, became the highest-grossing R-rated film of all time.  } \\
\multicolumn{2}{l}{ \textbf{NS Pretrained:} ryan reynolds teased his upcoming movie as the character of deadpool .} \\
\multicolumn{2}{l}{ \textbf{Standard Finetuned:} it 's a popular movie . } \\
\multicolumn{2}{l}{ \textbf{WD($\theta_\text{pre}$):} yes , i really like him . he is a very funny character .} \\
\multicolumn{2}{l}{ \textbf{Mix-review:} ryan reynolds . } \\
\hline
\end{tabular}
\caption{Templates for news or dialogue-style triggers. ``X'' is to be replaced by specific knowledge terms. They are followed by reference description and model samples for ``pokemon'' and ``deadpool''. Note that the pretrained model's sample is from news-style triggers, and the other samples are from dialogue-style triggers.}
\label{tab:knowledge_example_trigger}
\end{table*}

To query whether the model has knowledge of a certain term, we design three news-style and three dialogue-style ``trigger templates'' to trigger the model to generate responses related to the knowledge term. We collect 10 samples for each trigger, then we compute BLEU score of generated samples against the reference descriptions. We show some examples of trigger inputs in Table \ref{tab:knowledge_example_trigger}.

\begin{table}[h]
\small
\addtolength{\tabcolsep}{-1.1pt}
\begin{tabular}{|l|c|c|c|}
\hline
  \textbf{Training}   & \textbf{Dailydialog} & \textbf{Switchboard} & \textbf{Cornell}     \\ \hline
Pretrained & \multicolumn{3}{c|}{\bf BLEU-2 0.347 / BLEU-3 0.153}          \\ \hline
Baseline      & 0.124/0.007   & 0.032/0.003 & 0.081/0.003 \\ \hline
NS+finetune   & 0.162/0.047   & 0.187/0.052 & 0.207/0.071 \\ \hline
NS+WD           & 0.226/0.080   & 0.203/0.070 & 0.285/0.114 \\ \hline
NS+MR   & {\bf 0.261/0.108}   & {\bf 0.223/0.079} & {\bf 0.396/0.190} \\ \hline
\end{tabular}
\caption{Average BLEU-2/BLEU-3 scores for the model's samples w.r.t. the reference description. We highlight the pretrained model's performance for news triggers and the performance of the best model finetuned with dialogue data for dialogue triggers.}
\label{tab:main_knowledge}
\end{table}

The BLEU scores are shown in Table \ref{tab:main_knowledge}. Note that for the pretrained model we feed news triggers, while for the other dialogue models dialogue triggers are used. We observe that although the finetuned model is more knowledgeable than the baseline model, its score is much lower than the pretrained model. This demonstrates the forgetting problem for the standard finetuning.

On the other hand, we find that mix-review and WD($\theta_\text{pre}$) can effectively retain the knowledge acquired during pretraining, giving a much higher BLEU score than the standard finetuned model. Mix-review shows higher BLEU scores than WD($\theta_\text{pre}$), demonstrating its superiority in facilitating knowledge retention. We showcase samples from different models in Table \ref{tab:knowledge_example_trigger}. To save space, we manually select and show the most related sample out of the 30 samples for each knowledge term. The observations agree with the quantitative results: The standard finetuning loses the detailed information about the knowledge term, and mix-review helps the model retain it. More importantly, the model is able to express the knowledge in a dialogue context.

\begin{table}[]
\centering
\begin{small}
\begin{tabular}{|l|c|c|c|}
\hline
\textbf{Training} & \textbf{Knowledge}     & \textbf{Consistency}    & \textbf{Engaging}   \\ \hline
finetune   & $2.82 \pm .06$  & $4.28 \pm .06$ & $3.84 \pm .05$ \\ \hline
WD($\theta_\text{pre}$)         & $3.18 \pm .06$ & $4.60 \pm .06$ & $4.18 \pm .05$ \\ \hline
{Mix-review}  &  \textbf{3.40 $\pm$ .05} & {\bf 4.75 $\pm$ .06} & {\bf 4.27 $\pm$ .05 }  \\ \hline
\end{tabular}
\vspace{-0.2cm}
\caption{AMT rating scores (mean and standard deviation) for multi-turn knowledgeable dialogue evaluation.}
\vspace{-0.3cm}
\label{tab:multiturn_knowledge}
\end{small}
\end{table}

To further investigate our model's ability to conduct knowledgeable dialogues with users, we use the ParlAI\footnote{https://parl.ai/} platform to conduct multi-turn dialogue evaluation. For each session, the user will be assigned a random knowledge term from Google Trend, and have a 8-turn dialogue with the model under that topic. Ratings from around 600 dialogues are collected for each model, and are reported in Table \ref{tab:multiturn_knowledge}. In this evaluation we use the models finetuned on the Dailydialog data, because the nature of that dataset is closet to online chit-chat. It is observed that the model trained by mix-review significantly outperforms WD($\theta_\text{pre}$) on knowledge, consistency and engagingness, which agrees well with the results in Table \ref{tab:main_knowledge}. Some dialogue examples are included in Table \ref{tab:multiturn_example}.


\subsection{Behavior Analysis: Context Sensitivity}
\label{sec:bea_sens}

\begin{table}[]
\small
\centering
\begin{tabular}{|l|c|c|}
\hline
   \textbf{Training}          & \textbf{Dailydialog} & \textbf{Switchboard}      \\ \hline
NS Pretrained & \multicolumn{2}{c|}{\bf word shuffle +110\% / drop +105\%}        \\ \hline
Baseline      & +12\%/+28\%   & +4\%/+5\%      \\ \hline
MASS+FT & +24\%/+48\% & +15\%/+19\%   \\ \hline
NS+FT  & +41\%/+64\%   & +17\%/+21\%  \\ \hline
NS+WD($\theta_\text{pre}$)           & +26\%/+46\%   & +19\%/+25\%  \\ \hline
NS+MR   & {\bf +60\%/+108\%}  & {\bf +19\%/+30\%}  \\ \hline
\end{tabular}
\caption{The model's relative PPL drop when word-shuffle/drop is applied to input. ``FT'' refers to ``finetune''.  To save space, the results on the Cornell Movie dataset is deferred to Appendix \ref{appsec:expres}, Table \ref{tab:wordshuffle_app}. \label{tab:wordshuffle_main}}
\end{table}

The sensitivity to context is an important property for NLG models. However, as shown by \cite{hisotrydia19Chinnadhurai}, dialogue models trained from scratch typically are not sensitive to artificial distortion in the context input, showing the models have poor utilization of dialogue context. In this section, we repeat their experiments with pretrained or finetuned models. 

Following \cite{hisotrydia19Chinnadhurai}, we use two methods to distort the context input:
\begin{itemize}
    \item \textbf{word-drop:} We randomly drop 30\% of the words in the context input.
    \item \textbf{word-shuffle:} We randomly shuffle the words in the context input.
\end{itemize}

We use the relative drop in test-set perplexity to quantify the sensitivity. The results are presented in Table \ref{tab:wordshuffle_main}, where the result of the pretrained model is also included. First, we observe the baseline model trained from scratch is relatively insensitive to context, which agrees well with \citet{hisotrydia19Chinnadhurai}. The model with the standard pretrain-finetune process is much more sensitive, showing that pretraining effectively changes the model's behavior. Comparing to MASS, the NS pretrained model has better utilization of context, which explains its superior performance in PPL. 

Somewhat surprisingly, the finetuned dialogue models are much less sensitive to context input than the pretrained model without finetuning. This again verifies our worry in Section \ref{sec:forget_mixreview} that the model is forgetting some important generation skill during standard finetuning. Further, we find that the mix-review finetuning strategy can effectively alleviate this problem: Its sensitivity is much greater than that of standard finetuning, and is close to the pretrained model. 

\subsection{Behavior Analysis: Function Space Projection}
\label{sec:umap}

\begin{figure}
  \centering
  \includegraphics[width=\linewidth]{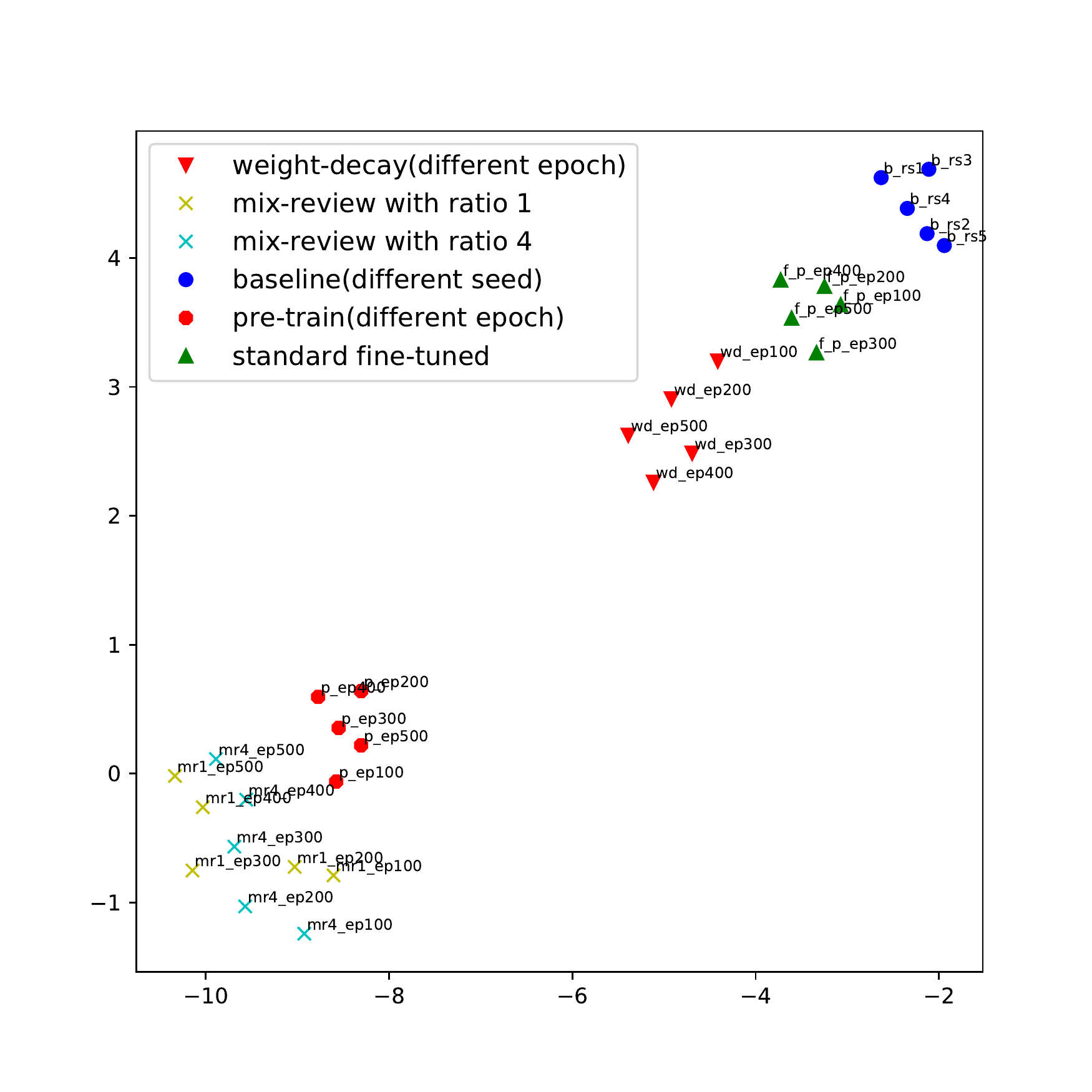}
  \caption{UMAP projection of checkpoints from different training processes.}
  \label{fig:100moutvec_umap}
\end{figure}

\begin{figure}
    \centering
  \includegraphics[width=\linewidth]{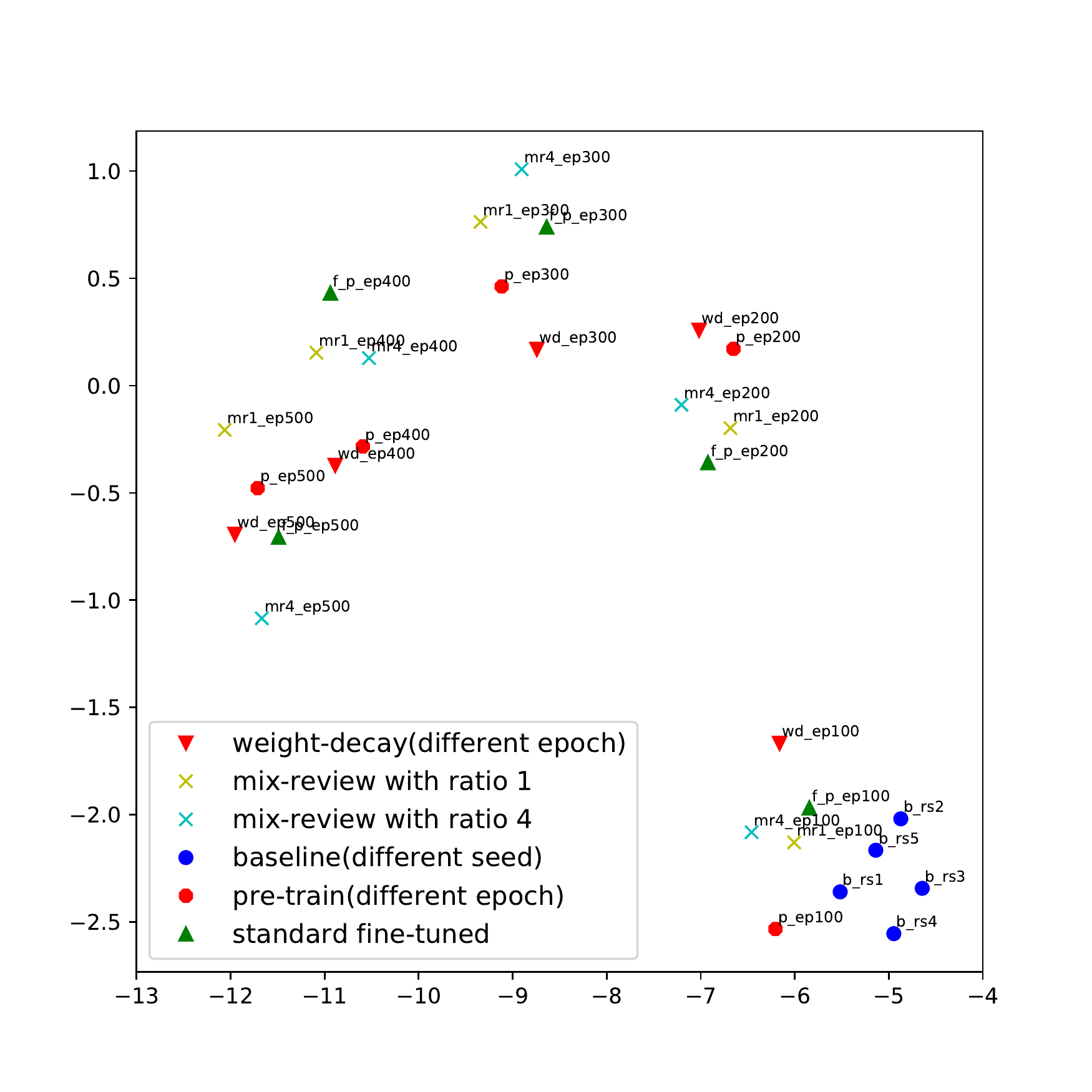}
  \caption{Parameter-space UMAP projection of checkpoints from different training processes.}
  \label{fig:100param_umap}
\end{figure}

It is interesting to study models' behavior via function-space 2D projection \citep{erhan10whypretrainhelp}.  We collect the model's output distribution on 10k words for the CCNEWS validation set and the Dailydialog validation set\footnote{It's a concatenation of two long vectors.}. And feed them as input to UMAP \cite{mcinnes2018umap-software}. We use the default hyper-parameter setting of the python implementation of UMAP.  The result is shown in Figure \ref{fig:100moutvec_umap}. Note that during pretraining of the CCNEWS data, 20 epochs are one entire data pass. We finetune from epoch 100, 200, 300, 400, 500 of the pretraining checkpoints.

We observe that the standard finetuned models are not close to the cluster of the pretrained models, which suggests the models' generative behavior is substantially different from the pretrained ones. Mix-review regularizes the finetuning process by keeping the model's generation behavior close to the pretrained model. These observations agree with our results in Section~\ref{sec:behav_knowledge} and \ref{sec:bea_sens}. Figure \ref{fig:100moutvec_umap} also suggests potential limitations of mix-review and WD($\theta_\text{pre}$): mix-review could be too ``aggressive'' and does not put enough attention on the target task. On the other hand, WD($\theta_\text{pre}$) is not strong enough in regularizing the model's generative behavior.

In Figure \ref{fig:100param_umap} we show the parameter-space UMAP projection for the same set of models. In this case, the input to UMAP is the concatenation of flattened weight matrices of the transformer model. A key observation is that the finetuned models are typically very close to the starting point (pretrained models). However, as shown in Figure \ref{fig:100moutvec_umap}, their behavior is very different. This suggests that a parameter-space regularization such as WD($\theta_\text{pre}$) could be not very effective for regularizing the model's behavior.


\section{Implications and Discussion}
\label{sec:implication}

\begin{table*}[]
\centering
\small
\begin{tabular}{l|l|l}
\hline
\begin{tabular}[c]{@{}l@{}} {\bf User:} have you been to china ?\\ {\bf Model:} yes , i have .\\ {\bf {\bf User:}} which city did you go ?\\ {\bf Model:} shanghai\\ {\bf {\bf User:}} how do you like it there ?\\ {\bf Model:} i like it very much\\ {\bf User:} i think the weather is not so good .\\ {\bf Model:} but it is nice in spring and fall .\end{tabular} &

\begin{tabular}[c]{@{}l@{}}{\bf U:} who is mark zukerberg ?\\ {\bf M:} he is the ceo of facebook .\\ {\bf U:} what is that company about ?\\ {\bf M:} it is a social media company .\\ {\bf U:} do you like him ?\\ {\bf M:} i think he is a good leader .\\ {\bf U:} my friend works at facebook .\\ {\bf M:} what does your friend do there ?\end{tabular} &

\begin{tabular}[c]{@{}l@{}}{\bf U:} tell me about tv show Robot Wars .\\ {\bf M:}  it ’s about robots fighting in a war .\\ {\bf U:} cool ! What season is it on ? \\ {\bf M:}  it ’s on next month .\\ {\bf U:} how many years has it been on for ? \\ {\bf M:} it started in 2003 . \\ {\bf U:} what channel does it come on ?\\ {\bf M:} the science channel. \end{tabular} \\ \hline

\multicolumn{3}{l}{\begin{tabular}[c]{@{}l@{}}{\bf User:} when is the independence day ? {\bf Best-of-10:} it 's july fourth.\\ {\bf User:} is coffee good for health ? {\bf Best-of-10:} it may help reduce your risk of certain diseases.\\ {\bf User:} should i do a phd or go to work ? {\bf Best-of-10:} a phd is not enough if you want to learn something new .\end{tabular}}  \\ \hline

\multicolumn{3}{l}{\begin{tabular}[c]{@{}l@{}}{\bf User:} oh shut up you . {\bf Worst-of-10:} shut up !\\ {\bf User:} you 're the problem .  {\bf Worst-of-10:} i hate you .\\ {\bf User:} i will kill . {\bf Worst-of-10:} i will kill you .\end{tabular}}                                                                           \\ \hline
\end{tabular}
\vspace{-0.1cm}
\caption{Multi-turn and single-turn examples of the model trained by mix-review on the Dailydialogue dataset. The single-turn examples involve light cherry-picking.}
\label{tab:multiturn_example}
\end{table*}

The sensitivity to dialogue context and the ability to transfer knowledge from pretraining opens the possibility of a \textbf{data-driven} knowledgable chat-bot. In Table \ref{tab:multiturn_example}, we show multi-turn and single-turn interaction examples with the model trained by mix-review. For demonstration purpose, we manually select the most interesting response out of 10 samples from the model for the single-turn examples. We observe that the model is able to return interesting responses with the knowledge it acquired from pretraining. Interestingly, it has developed its own ``opinions'' and can give advice to the user.

Next, we discuss the malicious response problem for open-domain dialogue models.  As shown by \cite{he2018detecting}, it is relatively difficult to trigger the dialogue models trained from scratch to output malicious responses. However, as shown in Table \ref{tab:multiturn_example}, the pretrained models are easily triggered to respond in a malicious way when ``provoked''. This is because compared to the baseline models, the pretrained models are more sensitive to the contextual input, making them easier to  manipulate. This makes the malicious response problem a more relevant issue to solve \cite{negtrain19tianxing}.

Finally, we discuss some limitations of our work. First, the mix-review strategy we proposed is a simple and preliminary attempt to alleviate the forgetting, and its performance is far from perfect. As shown in Appendix \ref{appsec:samples}, in a lot of cases, the generation from mix-review is still boring or non-informative. Next, the three datasets considered in this work are open-domain dialogue datasets, and they are not knowledge-intensive. It would be interesting, as future work, to check the forgetting problem for knowledge-grounded datasets such as Topical-chat \citep{Gopalakrishnan2019}.


\vspace{-0.1cm}
\section{Related Works}
\label{sec:related}
\vspace{-0.1cm}

\paragraph{Behavior of pretrained NLG Models}
Recently, multiple works \citep{radford2019language,jiang2020know,roberts2020knowledge,talmor2019olmpics,trinh2019lmcommonsense} have reported that pre-trained language models (LM) have implicitly stored large amounts of ``world knowledge'' in its parameters, and are able to answer common-sense questions. However, whether the world knowledge is well preserved after finetuning on target task dataset is not discussed.

On the other hand, knowledge-grounded NLG model \cite{liu18knowledgedialogue,guu2020realm,ijcai2018-commonsensedialogue} has been an important and exciting research topic. These studies usually involve additional retrieval modules or external knowledge bases to provide the model with relevant information. In contrast to these works, we study whether the model can conduct knowledgeable dialogues by itself.

\paragraph{Forgetting}
As discussed in Section \ref{sec:analyze_forgetting}, in contrast to the ``catastrophic forgetting'' problem in sequential learning \cite{pesudo18craig,Robins95catastrophicforgetting}, the performance drop on pretraining data is not necessarily bad for the NLP pretrain-finetune framework, and its implications have not been properly studied. In our analysis, we confirm the ``forgetting'' of important language generation skills during standard finetuning. The proposed mix-review strategy is similar to the \textit{pseudo-rehearsal} algorithm in sequential learning \cite{Robins95catastrophicforgetting}, with the difference being that we assume we still have access to the pretraining data. 




\section{Conclusion}

In this work, we attempt to answer to question of whether during finetuning, the model has forgotten some of the useful NLG skills acquired during large-scale pretraining. Through a set of detailed behavior analysis, we find the answer is, to some extent, yes. For example, the finetuned model fails to give detailed information about some knowledge terms, while the pretrained model can. As a preliminary attempt to alleviate the forgetting problem, we propose the mix-review finetuning method, and find it to be effective.

Our analysis shows that under the surface of the performance boost for standard metrics, large-scale pretraining changes the model's generative behavior in various profound ways. More importantly, the behavior change is influenced by the nature of data itself. For example, we demonstrate that we can discuss news with the dialogue model finetuned by mix-review, even when the target dataset is not about news (Dailydialog). We believe that this opens the possibility of a completely data-driven way to customize a language generator.

\bibliography{emnlp2020}
\bibliographystyle{acl_natbib}

\clearpage

\appendix

\section{Beam-search vs. Top-$k$ Sampling}
\label{appsec:beamsearch}

\begin{table*}[h]
\centering
\begin{tabular}{|l|c|c|c|c|}
\hline
              & \multicolumn{2}{c|}{\textbf{Beam Search}} & \multicolumn{2}{c|}{\textbf{Top-30 Sampling}} \\ \hline
Dataset      & \textbf{Entropy}   & \textbf{Max-ratio}   & \textbf{Entropy}     & \textbf{Max-ratio}     \\ \hline
Dailydialogue & 7.44 8.49          & 1.7\% 1.3\%          & 9.04 10.81           & 0.6\% 0.4\%            \\ \hline
Switchboard   & 4.96 5.54          & 34.9\% 27.8\%        & 8.47 10.45           & 8.4\% 7.9\%            \\ \hline
Cornell       & 6.10 6.56          & 10.2\% 9.9\%         & 8.76 10.54           & 1.4\% 1.1\%            \\ \hline
\end{tabular}
\caption{Average of diversity metrics for models on the three dialogue datasets.}
\label{tab:diversity_beamsearch}
\end{table*}

To compare beam-search with top-$k$ sampling (we set $k$ to 30), we compute diversity metrics for samples from models trained by different procedures (from scratch or pretrained). In particular, we compute bi-gram and tri-gram entropy, and the ratio of the most frequent response and second most frequent response (denoted as max-ratio) \citep{negtrain19tianxing}. The results are shown in Table \ref{tab:diversity_beamsearch}.

We observe that the responses given by top-$k$ sampling are much more diverse than beam-search. Beam-search suffers much from the ``generic response'' problem \citep{diversityjiwei16}, for example, 34\% of the responses are ``\texttt{um - hum}'' for Switchboard. Further, in our multi-turn dialogue experiments, beam-search is likely to give repetitive responses.  Finally, by manual inspection, we find the sample quality of top-$k$ sampling is not compromised. Due to these observations, we adopt top-$k$ sampling for our models.

\section{Details on Datasets}
\label{appsec:dataset}
Dailydialogue \citep{dailydialog17yanran} is a high-quality multi-turn dialog dataset. The language is human-written and less noisy. The dialogues in the dataset reflect our everyday communication and cover various topics about our daily life. The training split has around 11k dialogues (1.3 million words), and both the validation and test splits have 1k dialogues (0.1 million words).

The Switchboard Dialogue Act Corpus\footnote{\url{http://compprag.christopherpotts.net/swda.html}} is a version of the Switchboard Telephone Speech Corpus, which is a collection of two-sided telephone conversations, annotated with utterance-level dialogue acts. In this work we only use the conversation text part of the data, and select 1.1k dialogues for training (181k sentences / 1.2 million words), 50 dialogues for validation and 50 dialogues for testing. 

The Cornell Movie Dialogue Corpus\footnote{\url{https://www.cs.cornell.edu/~cristian/Cornell_Movie-Dialogs_Corpus.html}} \citep{cornell11cristian} is a collection of movie scripts. In the processing of the data, we simply regard the whole scripts from a movie as a long dialogue. The training split contains 9k dialogues (4.5 million words), and both the validation and test splits have 180 dialogues (85k words).

\section{Model Samples}
\label{appsec:samples}

In Table \ref{tab:appmodelsamples}, we show more samples from different training procedure, for the three dialogue datasets.

\begin{table*}
\vskip 0.15in
\begin{center}
\small
\begin{tabular}{l}
\hline
Dailydialogue \\
\hline
{\bf Input:} why do you think you are qualified for this position ? \\
\quad {\bf Baseline:} i can tell you more about your company because your company has a working experience \\ \quad with a company and company company . \\
\quad {\bf NS:} i have been working with microsoft for four years . i can make good use of microsoft ' s word pad \\ \quad and excel excel spreadsheet . \\
\quad {\bf MASS:} i think i have a strong sense of responsibility and i have had a high iq score in college . \\ \quad i am applying for some courses of study for some courses if i have a preference .  \\
\quad {\bf Mix-review:}  my background is in computer engineering and i have a strong commercial acumen in \\ \quad business - related application . i am in charge of all the things i like to do here .  \\
\quad {\bf WD($\theta_\text{pre}$):}  my background , education , experience as an english professor , and the work ethic are strong . \\
{\bf Input:}  according to your advertisement , you want an experienced software engineer . i think my \\ background meets the requirement of this position . \\
\quad {\bf Baseline:} yes . and you also need to know how to enjoy the work . and ,  here is a position to develop \\ \quad the company position under the company . \\
\quad {\bf NS:} do you know your qualifications ? \\
\quad {\bf MASS:} that ' s a good answer . i can tell you if we have made a decision this afternoon . \\
\quad {\bf Mix-review:}  i see . do you have some experience as an executive engineer ? \\
\quad {\bf WD($\theta_\text{pre}$):}  in your first year of operating your company , you had an impressive \\ \quad track record of product design , and achieved a high level of technical proficiency . \\
\hline
Switchboard \\
\hline
{\bf Input:} that 's going to be a really viable program if it continues to be funded at current levels\\
\quad {\bf Baseline:} oh really \\
\quad {\bf NS:} um it 's going to be a lot of fun it certainly is \\
\quad {\bf MASS:} sure sure sure \\
\quad {\bf Mix-review:}  well that that 's what i 'm saying too that 's that 's not gonna be an issue \\
\quad {\bf WD($\theta_\text{pre}$):}  yeah oh sure sure \\
{\bf Input:} uh - huh \\
\quad {\bf Baseline:} oh okay \\
\quad {\bf NS:} and uh i- i think there are some areas that if the government decides to close down that \\ \quad could be a real problem \\
\quad {\bf MASS:} and i really feel strongly that we don 't need the space station because we think there 's \\ \quad really no space \\
\quad {\bf Mix-review:}  but uh that \\
\quad {\bf WD($\theta_\text{pre}$):}  i mean it 's pretty i think it 's it 's pretty interesting to me that it 's uh \\
\hline
Cornell Movie\\
\hline
{\bf Input:} i expect you shortly . \\
\quad {\bf Baseline:} how so ? \\
\quad {\bf NS:} i 'll make a reservation . \\
\quad {\bf MASS:} i 'll come with you , doctor . \\
\quad {\bf Mix-review:}  no , i 'm fine \\
\quad {\bf WD($\theta_\text{pre}$):}  okay . \\
{\bf Input:} what shall we do ? \\
\quad {\bf Baseline:} we all don 't know where the hell he is now , but if he talks to me first , i 'll be back . \\
\quad {\bf NS:} nothing . \\
\quad {\bf MASS:} we send a cab ! i said i expect you to be fast . \\
\quad {\bf Mix-review:}  david 's not very lucid , but he 's always lucid . \\
\quad {\bf WD($\theta_\text{pre}$):}  stop him . \\
{\bf Input:} tea would be nice . \\
\quad {\bf Baseline:} i have no desire to go to bed . \\
\quad {\bf NS:} but you can 't come here and have tea too ? \\
\quad {\bf MASS:} i 've been expecting you , doctor . \\
\quad {\bf Mix-review:}  do you want me to drink some ? \\
\quad {\bf WD($\theta_\text{pre}$):}  what is it ? \\
\hline
\end{tabular}
\caption{Samples of different models on the Dailydialog/Switchboard/Cornell-Movie test-set. The samples are consecutive (input of the next sample is the reference response for the previous one).}
\label{tab:appmodelsamples}
\end{center}
\vskip -0.1in
\end{table*}

\section{Supplementary Experiment Results}
\label{appsec:expres}

\begin{table}[H]
\small
\centering
\begin{tabular}{|l|c|c|}
\hline
               & \multicolumn{2}{c|}{\bf Cornell}   \\ \hline
\textbf{ Model}          & \textbf{ Dialogue Triggers} & \textbf{ News Triggers} \\ \hline
{ NS Pretrained} & 0.245/0.089         & {\bf 0.347/0.153}          \\ \hline 
{ Baseline}       &    0.081/0.003      &   0.088/0.003   \\ \hline
{ NS+finetune}   &   0.207/0.071  &  0.207/0.063    \\ \hline
{ NS+WD($\theta_\text{pre}$)} & 0.285/0.114 & 0.202/0.072 \\ \hline
{ NS+Mix-review}    & {\bf 0.396/0.190}         &   0.212/0.065  \\ \hline
\end{tabular}
\caption{Average BLEU-2/BLEU-3 scores for the model's samples w.r.t. the reference description. We highlight the pretrained model's performance for news triggers and the performance of the best model finetuned with dialogue data for dialogue triggers.}
\label{tab:knowledge_app}
\end{table}

In this section we supplement results that are deferred in the main body due to space limit. 


In Table \ref{tab:knowledge_app} we show the knowledge transfer results for the Cornell Movie dataset.

In Table \ref{tab:wordshuffle_app} we show context sensitivity results for the Cornell Movie dataset.

\begin{table*}[]
\centering
\begin{tabular}{|l|l|l|l|}
\hline
{ Model(Dataset)}              & PPL({\bf normal}) & PPL({\bf word-shuffle}) & PPL({\bf word-drop}) \\ \hline
{ NS pretrained(CCNEWS)}          & 17.33    & 36.56({\bf +110.96\%})   & 35.56({\bf +105.19\%})  \\ \hline \hline
{ Baseline(Cornell)}      & 49.48  & 50.22(+1.4\%)     & 50.85(+2.7\%)  \\ \hline
{ MASS+finetune(Cornell)}          & 30.25 & 36.50(+20.6\%) & 36.36(+20.1\%)  \\ \hline
{ NS+finetune(Cornell)}            & 28.06  & 36.88(+31.4\%) & 34.47(+22.8\%)   \\ \hline
{ NS+WD($\theta_\text{pre}$)(Cornell)} & 27.80 & 37.46({\bf +34.7\%}) & 35.10({+26.2\%}) \\ \hline
{ NS+Mix-review(Cornell)} &  27.54  & 36.94({ +34.1\%}) & 37.72({\bf +36.9\%})  \\ \hline
\end{tabular}
\caption{The model's PPL performance when word-shuffle or word-drop is applied to the context input. On the left we describe what training process is used and on which test set is PPL evaluated. Note that MASS/NS refers to MASS/NS pretraining with standard finetuning.}
\label{tab:wordshuffle_app}
\end{table*}

\end{document}